\documentclass[letterpaper]{article} % DO NOT CHANGE THIS
\usepackage{aaai23}  % DO NOT CHANGE THIS
\usepackage{times}  % DO NOT CHANGE THIS
\usepackage{helvet}  % DO NOT CHANGE THIS
\usepackage{courier}  % DO NOT CHANGE THIS
\usepackage[hyphens]{url}  % DO NOT CHANGE THIS
\usepackage{graphicx} % DO NOT CHANGE THIS
\usepackage{natbib}  % DO NOT CHANGE THIS AND DO NOT ADD ANY OPTIONS TO IT
\usepackage{caption} % DO NOT CHANGE THIS AND DO NOT ADD ANY OPTIONS TO IT
\usepackage{algorithm}
\usepackage{algorithmic}

\usepackage{newfloat}
\usepackage{listings}
\DeclareCaptionStyle{ruled}{labelfont=normalfont,labelsep=colon,strut=off} % DO NOT CHANGE THIS
\floatstyle{ruled}
\newfloat{listing}{tb}{lst}{}
\floatname{listing}{Listing}
\nocopyright 
\usepackage{booktabs}
\usepackage{amsmath,amssymb} % define this before the line numbering.

\newcommand{\eg}{\emph{e.g.}}
\newcommand{\ie}{\emph{i.e.}}

\newcommand{\comp}[1]{{#1}}

\title{Audio-Visual Contrastive Learning with Temporal Self-Supervision} % Replace with your title
\author {
    Simon Jenni,\textsuperscript{\rm 1}
    Alexander Black, \textsuperscript{\rm 2}
    John Collomosse \textsuperscript{\rm 1,\rm 2}
}
\affiliations {
    \textsuperscript{\rm 1} Adobe Research \\
    \textsuperscript{\rm 2} University of Surrey\\
    jenni@adobe.com, alex.black@surrey.ac.uk, collomos@adobe.com
}
\usepackage{bibentry}
\begin{document}

\maketitle

\begin{abstract}
We propose a self-supervised learning approach for videos that learns representations of both the RGB frames and the accompanying audio without human supervision.   
In contrast to images that capture the static scene appearance, videos also contain sound and temporal scene dynamics.  
% To effectively capture the aural and temporal features in video we propose a model that combines multi-modal contrastive learning with temporal supervision. 
% Our method integrates both temporal self-supervision and intra- and inter-modal contrastive learning to leverage the temporal and aural dimension inherent to videos.
To leverage the temporal and aural dimension inherent to videos, our method extends temporal self-supervision to the audio-visual setting and integrates it with multi-modal contrastive objectives.
% Our method extends temporal self-supervision to the audio-visual setting and integrates it with multi-modal contrastive objectives to capture the temporal and aural dimensions inherent to videos.
% As temporal self-supervision, we consider playback-speed and -direction recognition in both modalities and intra- and inter-modal temporal ordering tasks. 
As temporal self-supervision, we pose playback speed and direction recognition in both modalities and propose intra- and inter-modal temporal ordering tasks. 
% Furthermore, we propose a novel contrastive objective in which the usual positive pairs obtained through data augmentation are supplemented with additional positives and negatives sampled from the evolving feature space. 
Furthermore, we design a novel contrastive objective in which the usual pairs are supplemented with additional sample-dependent positives and negatives sampled from the evolving feature space. 
In our model, we apply such losses among video clips and between videos and their temporally corresponding audio clips. 
% We then apply such losses among temporally augmented video clips and between videos and their temporally aligned audio clips while excluding any contrastive terms among audio clips. 
% We then apply such losses among temporally augmented video clips and between videos and their temporally aligned audio clips while excluding any contrastive terms among audio clips. 
We verify our model design in extensive ablation experiments and evaluate the video and audio representations in transfer experiments to action recognition and retrieval on UCF101 and HMBD51, audio classification on ESC50, and robust video fingerprinting on VGG-Sound, with state-of-the-art results. 

\end{abstract}

\section{Introduction}

Videos provide a rich source of information for audio-visual learning.
Besides static moments in time (single video frames), they also contain the scene dynamics (object motion) and often include the sounds of the environment and scene objects.
It seems hopeless to learn general representations that capture this rich semantic information in videos, \ie, their appearance, motions, and sounds from such high-dimensional data through sparse human supervision. 
Self-supervised learning (SSL) \cite{Carl2015,chen2020simple,he2020momentum} has emerged as a viable alternative to supervised learning in recent years.
Such methods might be better suited for general video representation learning since they are not constrained by the prohibitive cost of exhaustive human annotations on video. 
However, since most current self-supervised methods are tailored to static images, they might not effectively use videos' added temporal and aural dimensions. 
A self-supervised learning task that successfully integrates the static scene appearance and the aural and temporal features potentially results in a representation that better generalizes to downstream vision applications, such as action recognition, video retrieval, or robust video content fingerprinting.

Indeed, recent works that explored the aural and temporal dimensions of videos in isolation have demonstrated that they are both effective self-supervision signals. 
Several works \cite{Morgado_2021_CVPR,patrick2020multi,alwassel2019self} demonstrate that audio-visual contrastive learning often performs better than uni-modal contrastive learning (\ie, using only the RGB frames). 
Likewise, temporal reasoning tasks \cite{misra2016shuffle,jenni2021time,dave2021tclr} have demonstrated good transfer performance, especially for downstream tasks where motion is the main discerning factor (as opposed to static scene appearance).

\begin{figure}[t]
    \centering
    \includegraphics[width=\linewidth]{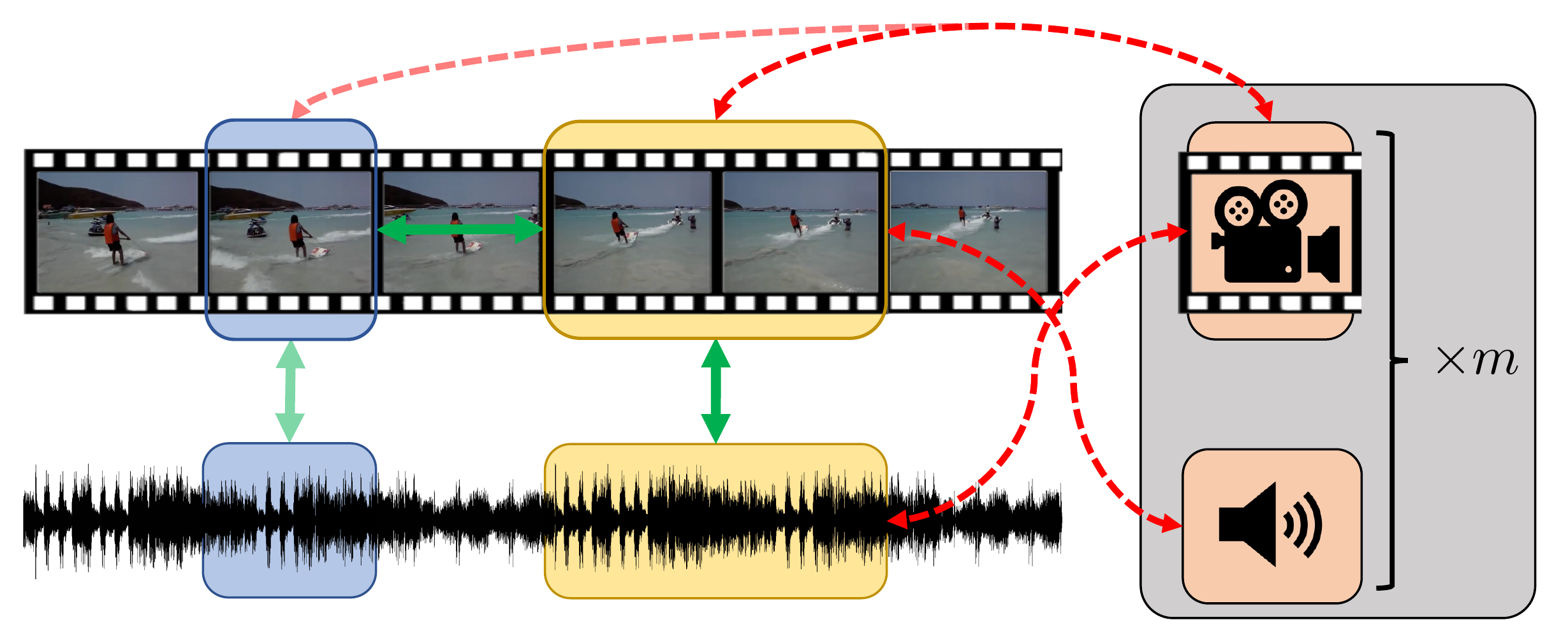}

    \caption{\textbf{Illustration of Contrastive Loss Terms in our Model.} 
    We demonstrate the main contrastive pairs in our formulation given an example video clip (yellow box in the middle) and its corresponding audio clip. Positives (solid green arrows) are constructed from differently augmented video clips of the same training instance (blue box) and \emph{temporally aligned} pairs of the corresponding video and audio clips.
    Negatives (dashed red arrows) stem from $m$ other video and audio clips from the current mini-batch or a memory bank of prior embeddings (gray box on the right). 
    Additional positives from the memory bank are omitted from the figure.  
    Note that our formulation does not contain any contrastive terms among audio clips. 
    }
    \label{fig:contrastive_terms}
\end{figure}

\begin{figure}[t]
    \centering
    \includegraphics[width=\linewidth]{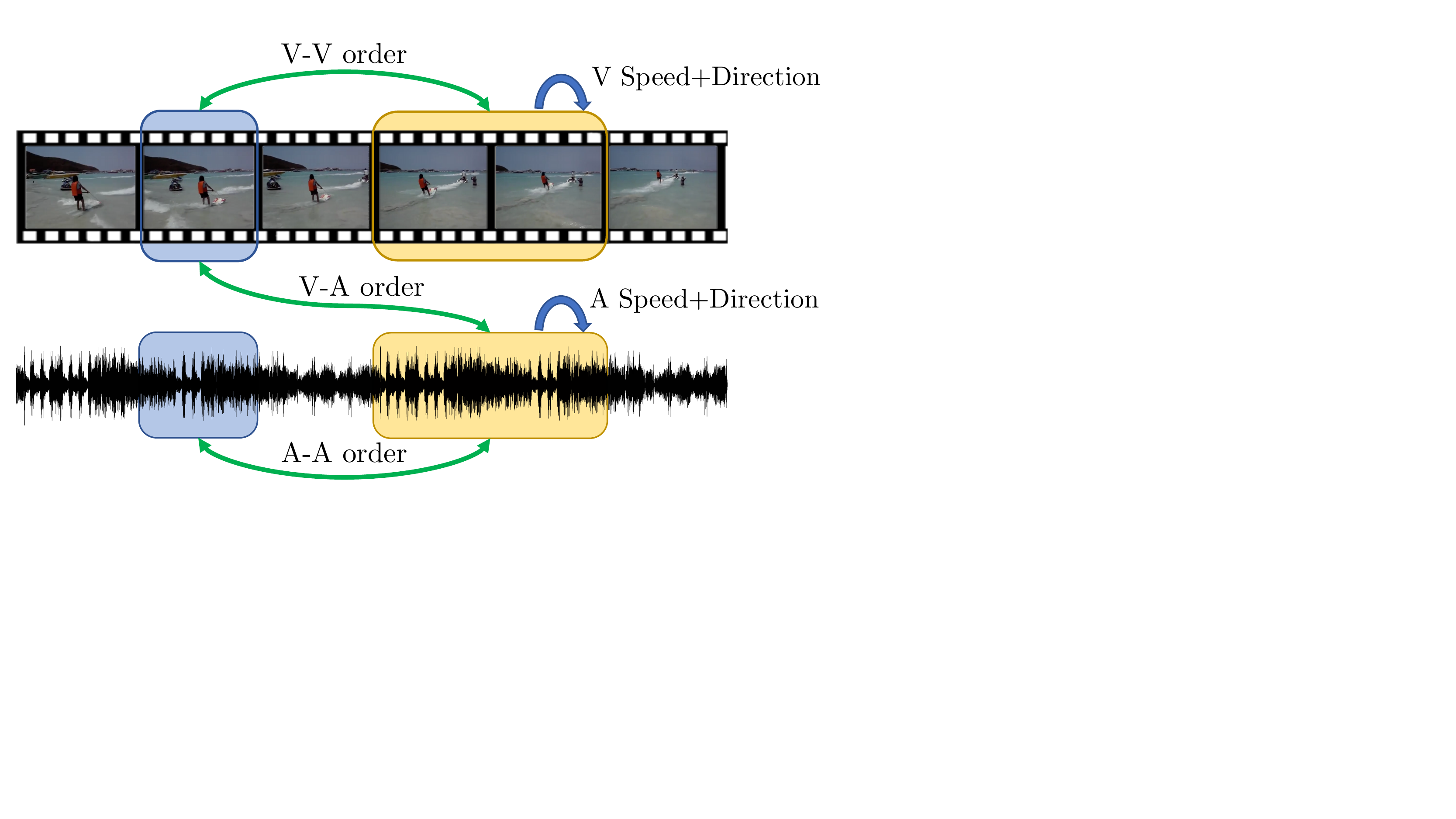}
    \caption{\textbf{Illustration of the Temporal Reasoning Tasks.}
    Besides contrastive terms, our model encompasses both per-clip classification tasks (blue arrows) about the playback-speed and -direction, and temporal ordering tasks (green arrows) which are performed both intra- and cross-modal (V: RGB frames, A: audio). 
    }
    \label{fig:temp_ssl}
\end{figure}

% In contrast, our work aims to leverage both dimensions (sound and time) as learning signals in a unified model architecture and training objective. 
In contrast, our work aims to leverage both sound and time as learning signals in a unified model architecture and training objective. 
To this end, we extend temporal self-supervision to the audio domain and propose cross-modal audio-visual temporal reasoning tasks. 
Concretely, we pose playback-speed and -direction recognition \cite{wei2018learning,benaim2020speednet,jenni2020video}, as a pretext task for audio representation learning and propose temporal clip ordering as a task for both intra-modal (\eg, audio-audio) and cross-modal (\eg, audio-video) learning (see Figure~\ref{fig:temp_ssl}). 
Furthermore, we introduce a model architecture and training objective for contrastive audio-visual learning that supplements these temporal learning tasks. 
Towards this goal, we carefully study how the inclusion and exclusion of different intra- and inter-modal contrastive objectives influences downstream performance. 
% Our key findings are 1. video-video contrastive terms helps 2. temporally aligned positives are best for cross-modal contrastive learning, and 3. audio-audio contrastive terms hurt downstream feature performance (see Figure~\ref{fig:contrastive_terms}). 
Our key findings for optimal audio-visual contrastive learning are 1. inclusion of video-video contrastive terms 2. temporally aligned cross-modal positives, and 3. exclusion of audio-audio contrastive terms (see Figure~\ref{fig:contrastive_terms}).
% Our design decisions are 1. inclusion of video-video contrastive terms 2. using temporally aligned cross-modal positives, and 3. exclusion of audio-audio contrastive terms (see Figure~\ref{fig:contrastive_terms}). 

We further explore the design of the contrastive loss terms \cite{wu2018unsupervised}, \ie, how to build positive and negative pairs for effective learning. 
In constructing our contrastive objective, we take inspiration from recent image-based methods \cite{dwibedi2021little,koohpayegani2021mean} and extend the set of positive samples with nearest neighbors in the evolving feature space. 
Thus, besides standard augmented views for positive sampling, we consider nearest neighbors sampled from a queue of prior embeddings as additional positives. 
Notably, the neighborhood structure and sample weights are both calculated through cross-view similarity, \ie, either through the feature space similarity to the augmented view (for intra-modal learning) or the temporally aligned sample from the other modality (for cross-modal learning). 
We also use this cross-view induced neighborhood structure to sample negative pairs in a sample-dependent manner.
This allows us to control the difficulty of the negative samples, \eg, preventing ambiguous or confusing negatives resulting from duplicates or heavy class imbalance, while also preventing possible collapse through the absence of any negatives.

We verify our model design in extensive ablation experiments and compare it to prior works in established action recognition and retrieval benchmarks on UCF101 and HMDB51. 
We also evaluate the audio branch of our model for environmental sound classification on ESC50.
Finally, we demonstrate the effectiveness of fusing the learned audio-visual features for downstream video classification on Kinetics-600 and VGG-Sound, and for robust video content retrieval under novel content manipulations for video fingerprinting \cite{lee2008robust,black2021vpn} on VGG-Sound.
% Finally, we evaluate our model for robust video content retrieval under novel content manipulations for video fingerprinting \cite{lee2008robust,black2021vpn} on VGG-Sound \cite{chen2020vggsound}. 
We investigate video fingerprinting as a novel downstream application due to its growing importance given the ever-expanding scale of visual data online and the increasing threat and sophistication of malicious content manipulations. 
\\

\noindent \textbf{Contributions.} To summarize, we make the following contributions: 1) We introduce temporal self-supervision in the audio domain and the cross-modal setting; 2) We propose a contrastive loss design that extends the usual contrastive pairs with sample-dependent positives and negatives; 
% 3) We propose a model that combines temporally aligned cross-modal and video-video contrastive objectives with temporal self-supervision;
3) We explore various multi-modal contrastive model designs and demonstrate the importance of a) using temporally aligned positives for cross-modal terms and b) excluding audio-audio contrastive terms;
4) Finally, we demonstrate the quality of the learned audio-visual features in extensive transfer experiments to action recognition, video retrieval, audio classification, and a novel video fingerprinting benchmark. 

\section{Prior Work}

\noindent \textbf{Contrastive Video Representation Learning.}
Contrastive learning is arguably the most popular self-supervised learning approach in computer vision today.
These methods are typically based on the task of discriminating training instances up to strong data-augmentation \cite{dosovitskiy2015discriminative,wu2018unsupervised}, which was shown to be remarkably effective for unsupervised image representation learning \cite{chen2020simple,he2020momentum} and has inspired a line of novel self-supervised methods \cite{grill2020bootstrap,chen2020exploring,caron2020unsupervised,wang2020unsupervised}.
Recently, methods were proposed that extend the set of positive pairs with nearest neighbors in the learned embedding space \cite{dwibedi2021little,koohpayegani2021mean}. Our loss design similarly uses the evolving feature space to extend the set of contrastive pairs. 
In contrast, our loss design retains the exact match, contains multiple positives weighted based on cross-view similarity, and uses additional sample-dependent negatives.

Several recent works have explored contrastive learning on video. 
When dealing with video, the set of data augmentations can be extended with several temporal augmentations (\eg, temporal crops). 
A natural extension is thus to add temporal augmentations to the set of data-augmentations that define the positive pairs for contrastive learning \cite{qian2020spatiotemporal,feichtenhofer2021large}.
Other works instead propose to learn to discriminate among temporally augmented clips \cite{dave2021tclr,patrick2020multi}, or learn to recognize the temporal input transformations in a multi-task approach \cite{bai2020can,jenni2021time}.
Our model combines contrastive learning among video clips with audio-visual contrastive and temporal self-supervised learning.

\noindent \textbf{Temporal Self-Supervision.}
Classic self-supervised approaches were based on so-called pretext tasks. 
On images, popular examples are the ordering of image patches \cite{Carl2015,noroozi2016unsupervised}, the colorization of gray-scale images \cite{zhang2016colorful,zhang2016split}, or the classification of sets of image transformations 
\cite{gidaris2018unsupervised,jenni2018artifacts,jenni2020steering}.
Pretext tasks that turned out particularly successful on video are based on recognizing temporal transformations. 
Some works explored the ordering of video frames \cite{misra2016shuffle,brattoli2017lstm,fernando2017self,lee2017unsupervised} or whole video clips \cite{xu2019self,kim2019self}, others the classification of the playback direction \cite{wei2018learning}, the playback speed \cite{epstein2020oops,benaim2020speednet,yao2020video}, or general temporal warpings \cite{jenni2020video}. 
We also leverage temporal supervision and extend it to multi-modal audio-visual representation learning.

\noindent \textbf{Audio-Visual Self-Supervised Learning.}
Another source of self-supervision on video can be found in the accompanying sound. 
Early works explored audio to learn single frame representations, \eg, by predicting summary statistics of the sounds corresponding to a frame \cite{owens2016ambient}, or by recognizing if an audio snippet and image are temporally aligned \cite{Owens_2018_ECCV,arandjelovic2017look}. 
Similar to these image-based approaches \cite{korbar2018cooperative} learned audio and video representations by recognizing when audio and video signals are synchronized.
More recently, contrastive audio-visual learning for video achieved remarkable performance \cite{recasens2021broaden}.
For example, \cite{alwassel2020self} performs clustering in one domain (\eg, audio) and uses the resulting clusters as supervision for the other domain (\eg, video). 
\cite{Morgado_2021_CVPR} demonstrate the effectiveness of cross-modal audio-visual contrastive learning and extend the set of positive samples within a modality with samples that show high cross-modal agreement. 
Other works even include language in audio-visual contrastive models \cite{alayrac2020self,akbari2021vatt}.
We instead focus on audio-visual learning and propose incorporating temporal supervision in both modalities.

\section{Model}

Let $\mathcal{D}_v=\left\{v_{1}, v_{2}, \ldots, v_{N}\right\}$ be a set of unlabelled training videos and let $\mathcal{D}_a=\left\{a_{1}, a_{2}, \ldots, a_{N}\right\}$ be their corresponding audio tracks. 
Our goal is to learn a video encoder $F_v$ (a 3D-ConvNet) and an audio encoder $F_a$ (a 2D-ConvNet) without human supervision. 
The inputs to the two networks are assumed to be of shape $v_i \in \mathbb{R}^{T\times H\times W \times C}$ and $a_i \in \mathbb{R}^{f\times t}$, where $a_i$ is a spectogram representation of the audio track.

\noindent \textbf{Temporal Input Augmentations.} 
An essential component of modern SSL approaches is the set of data augmentations applied to the input. 
In contrastive learning, these input transformations define the set of learned invariances. They typically comprise color jittering and geometric transformations, like random resizing, cropping, and horizontal flipping.
For our method, temporal transformations, \ie, random temporal cropping and manipulations of the playback speed and direction, are particularly important. We will thus indicate the precise temporal manipulations with $\tau_r$. 
Furthermore, we assume that $\tau_r$ has consistent behavior across modalities, \ie, $\tau_r(v_j)$ and $\tau_i(a_r)$ represent the exact same moments in time for the audio and video domain.

\subsection{Intra- and Inter-Modal Contrastive Learning}

Our training objective comprises several predictive-contrastive loss terms.
In general, we formulate these losses based on the two modalities involved and on the direction of the prediction, \eg, indicating that the visual representation is being predicted from the audio. 
For the purpose of this discussion let $\nu_i^r=\psi_v( F_v(\tau_r(v_i)))$ denote the output of the video encoder followed by a projection MLP $\psi_v$ for the input $\tau_r(v_i)$. Let similarly $\alpha_i^r=\psi_a( F_a(\tau_r(a_i)))$ be the feature vector for the corresponding audio track. 
Let further $\hat{\nu}_i^s$ be the feature of a different augmentation of the video $v_i$.

We illustrate the general form of the contrastive objective using the video-to-audio loss term, which is given by
% \begin{equation}
\begin{multline}
\textstyle
    \ell_{va}(\nu_i^r, \mathcal{P}_i^{va}, \mathcal{N}_i^{va})= \\ \sum_{p,w \in \mathcal{P}_i^{va}} -w \log \left( \frac{d( \phi_v(\nu_i^r) , p)}{d(\phi_v(\nu_i^r), p) +\sum_{n \in \mathcal{N}_i^{va}} d(\phi_v(\nu_i^r) , n)}\right),
    \label{eq:crl_loss}
% \end{equation}
\end{multline}
where $\phi_v$ denotes a predictor MLP (following prior work \cite{grill2020bootstrap}) and 
\begin{equation}
    d(x, y):= \exp \left(\frac{1}{\lambda}  \frac{ {x}^\intercal { y}}{\Vert {x} \Vert_2 \Vert {y} \Vert_2} \right),
    \label{eq:sim}
\end{equation}
is a measure of the similarity between the feature representations of $x$ and $y$, and $\lambda=0.2$ is a temperature parameter. 
Note that we do not back-propagate through the second argument $y$ in Eq.~\ref{eq:sim}.
In this general formulation, the set $\mathcal{P}_i$ defines the instance-dependent positive samples along with their weighting factor $w$, and $\mathcal{N}_i$ defines the negatives for contrastive learning.

\noindent \textbf{Sources for Positive and Negative Contrastive Samples. }
We consider two sources for sampling the positive and negative pairs of the contrastive loss terms: 1. the set of examples in the mini-batch $\mathcal{B}$ at each iteration, and 2. a memory bank of prior feature embeddings.  
In our model, we maintain a memory bank $Q_v$ (implemented as a FIFO queue) for the video domain and a corresponding $Q_a$ for audio. 
Let $|Q_v|=|Q_a|=n_q$ be the size of the memory banks and let $\text{NN}_{j:k}(\nu, Q_v)$ denote the sequence $\{\eta_j, \ldots, \eta_k\}$ from the $j$-th to the $k$-th nearest-neighbor of $\nu$ in $Q_v$.
For positive examples from the memory bank we further introduce a set of loss term weights $W_{1:k}(\nu):=\{\omega_1, \ldots, \omega_k \}$, where each weight is given by
\begin{equation}
    \omega_j:= \frac{d({\nu} , \eta_j)}{\sum_{\eta_l \in \text{NN}_{1:k}(\nu_i, Q_v)} d({\nu}, \eta_l)},
    \label{eq:weight}
\end{equation}
thus weighting each nearest neighbor proportional to their similarity to $\nu$.
The memory banks are updated with the mean of the features from the two augmented views in each mini-batch, \ie, $(\nu+\hat{\nu})/2$ in the case of $Q_v$.

We will now describe different instantiations of the contrastive losses and their positive and negative sample sets for the intra-modal and cross-modal objectives.

\noindent \textbf{Visual-Visual Contrastive Term $\ell_{vv}$.} 
For a video feature vector $\nu_i^r$ in the case of video-video contrastive learning, we set $\mathcal{P}_i^{vv}=\{(\hat{\nu}_i^s, 1)\} \cup \text{NN}_{1:k}(\hat{\nu}_i^s, Q_v) \times W_{1:k}(\hat{\nu}_i^s) $, where $\text{NN}_{1:k}(\hat{\nu}_i^s, Q_v)$ is the set of the first $k$ nearest neighbors of $\hat{\nu}_i^s$ extracted from $Q_v$. We set $k=5$ in our experiments.
The set of negatives is constructed as $\mathcal{N}_i^{vv}=\{\nu_j \in \mathcal{B} | j \neq i \}  \cup \text{NN}_{q:q+m}(\hat{\nu}_i^s, Q_v) $ and contains all the video features not belonging to $v_i$ that are in the current training mini-batch $\mathcal{B}$, as well as $m$ additional negatives sampled from the memory queue as the $q$-th up to the  $(q+m)$-th nearest neighbor of $\hat{\nu}_i^s$.
By default we set $q=\frac{n_q}{2}$, thus starting from the neighbor in $Q_v$ with median distance to $\hat{\nu}_i^s$ and set $m=2048$.

\noindent \textbf{Audio-Visual Contrastive Terms $\ell_{va}$ and $\ell_{av}$.} 
Since the terms $\ell_{va}$ and $\ell_{av}$ and the definition of their respective positive and negative sets is symmetric, we will restrict our illustration to the case of $\ell_{va}$.
Given a video feature vector $\nu_i^r$, we set $\mathcal{P}_i^{va}=\{({\alpha}_i^r, 1)\} \cup \text{NN}_k({\alpha}_i^r, Q_a)\times W_{1:k}({\alpha}_i^r) $, where ${\alpha}_i^r$ is the feature of the corresponding audio clip with \emph{identical temporal augmentation} (note the superscript).
This is in contrast to the definition of $\ell_{vv}$ where positive pairs were not temporally aligned. 
As we will show in ablations, we found temporal alignment to be important for cross-modal contrastive learning. 
The set of negatives is defined as $\mathcal{N}_i^{va}=\{\nu_j \in \mathcal{B} | j \neq i \} \cup \{\alpha_j \in \mathcal{B} | j \neq i \} \cup \text{NN}_{q:q+m}({\alpha}_i^r, Q_a) $, \ie, we consider both other audio and other video feature vectors as negatives.

\noindent \textbf{Multi-Modal Contrastive Objective.}
Our final contrastive objective is composed of the following intra- and inter-modal terms 
\begin{multline}
    \mathcal{L}_{\operatorname{CRL}}=\mathop{\mathbb{E}}_{v_i,a_i}[\ell_{vv}(\nu_i^r, \mathcal{P}_i^{vv}, \mathcal{N}_i^{vv}) + \\ \ell_{va}(\nu_i^r, \mathcal{P}_i^{va}, \mathcal{N}_i^{va}) + \ell_{av}(\alpha_i^r, \mathcal{P}_i^{av}, \mathcal{N}_i^{av}) ] .
    \label{eq:crl_obj}
\end{multline}
Note that our final model does not contain an audio-audio contrastive term. Indeed, we find that including such a term analogous to $\ell_{vv}$ hurts the final feature performance in transfer experiments (see ablations in Table~\ref{tab:mm_abl}).
An illustration of the intra- and inter-modal terms is given in Figrue~\ref{fig:contrastive_terms}.

\subsection{Temporal Self-Supervision for Video and Audio}\label{sec:temp}

Aside from learning from the correspondence between audio and video as proposed above, we also want to promote the learning of temporal features in both domains through self-supervised temporal reasoning tasks. 
These temporal pretext tasks can be categorized into unitary intra-modal tasks and pairwise intra- and cross-modal objectives (see Figure~\ref{fig:temp_ssl}).

\noindent \textbf{Intra-Modal Speed and Direction Classification.}
To capture short-term temporal video features we leverage the classification of temporal transformations as SSL objectives \cite{jenni2020video}. 
Concretely, we train the model to predict whether videos are played forward or backward and at which playback speed.
The direction classification is a simple binary classification task per clip, and either direction is equally likely during training. 
The speed classification is posed as a classification task among 4 speed classes ($1\times$, $2\times$, $4\times$, and $8\times$ speedup). The speed manipulations are implemented via temporal subsampling, and all the speed classes are equally likely during pre-training. 

We propose to leverage such temporal supervision in the audio domain in this work. 
We apply the temporal transformations to the 1D raw audio signal (analogous to the video domain), \ie, we subsample the signal for speed manipulations and reverse its direction before computing the spectrogram. 
In experiments, we also investigate an alternative approach where we perform the temporal transformations in the audio spectrogram (thus not manipulating the frequency). Interestingly, we found that transforming the raw audio waveform is much more effective, even when accounting for processing artifacts in manipulating the spectrogram (see ablations in Table~\ref{tab:audio_abl}). 

% The extension of the direction classification is straightforward and amounts to recognizing a horizontal (temporal) flipping in the audio spectrogram. 
% The implementation of the audio speed prediction requires more care. 
% An equivalent subsampling of the audio sound wave would result in a pitch shift of the signal, thus reflecting the manipulation in the time- and frequency domain of the spectrogram.
% To make the manipulation less obvious, we randomize the Short-Term-Fourier-Transform step size and resize the resulting spectrograms to a fixed size of $224\times224$. 
% This modification ensures that the temporal dimension of the spectrogram is consistent for all the speed classes (due to the resizing), prevents a pitch-shifting of the signal, and reduces shortcuts through processing artifacts (due to the random step-size). 

\noindent \textbf{Intra- and Inter-Modal Temporal Ordering.}
To capture the longer-term dynamics of videos we propose to also perform temporal learning tasks at the clip level by predicting the order of two video clips.   
% Reasoning about the temporal ordering of video clips has been shown to be a useful pretext task \cite{jenni2021time,xu2019self,kim2019self}.
Besides performing such temporal ordering solely on video \cite{jenni2021time,xu2019self,kim2019self}, we extend it to temporal ordering of the audio tracks and cross-modal audio-visual temporal ordering. 
Concretely, we pose the three-way classification of two temporal signals into 1. correctly ordered, 2. overlapping, and 3. wrongly ordered. 
This task is implemented by concatenating the representations of the two time-signals along the channel dimension and feeding it through a classifier, \eg, $\phi_{va}([F_v(v_i), F_a(a_i)])$ for video-audio ordering. 
Likewise, we introduce classifiers $\phi_{vv}$, $\phi_{av}$, and $\phi_{aa}$ for video-video, audio-video, and audio-audio temporal ordering.

Finally, we jointly optimize the network weights of the audio and video branch on the combination of the temporal and contrastive objectives. 
Concretely, let $\mathcal{L}_{\operatorname{TEMP}}=\mathcal{L}_{speed}+\mathcal{L}_{direction}+\mathcal{L}_{order}$ be the sum of all the losses for the above temporal reasoning tasks. 
The final objective is then given by
\begin{equation}
    \mathcal{L}_{\operatorname{SSL}} = \mathcal{L}_{\operatorname{CRL}} + \lambda \mathcal{L}_{\operatorname{TEMP}},
\end{equation}
where we set $\lambda=0.5$.

\subsection{Implementation Details}

For our video encoder $F_v$ we consider variants of the popular 3D-ConvNet architectures R3D \cite{hara2018can} and R(2+1)D \cite{tran2018closer}. 
If not specified otherwise, input video clips are assumed to contain 16 frames of resolution $112\times112$ for R(2+1)D, $128\times128$ for R3D-18,  and $224\times224$ for R3D-34. 
Our audio encoder $F_a$ is based on a standard ResNet-34 \cite{he2016deep} architecture in all experiments. 
Input spectrograms to the audio encoder are resized to $224\times224$. 

We train the models using the AdamW optimizer \cite{loshchilov2018decoupled} with a weight decay set to $10^{-4}$. 
The learning rate follows a cosine annealing schedule \cite{loshchilov2016sgdr} with a maximum learning rate of $3\cdot10^{-4}$ and linear warm-up in the first training epoch. 
By default, we train all the models with a batch size of 256. 

Besides the temporal input transformations described above (\ie, playback speed+direction changes and temporal cropping), we use the typical data augmentation recipe for contrastive methods, \ie, horizontal flipping, color-jittering, and random spatial cropping.
We do not apply any augmentations beyond the temporal ones for audio. 

The projection MLPs $\psi$ contain two hidden layers of size 1024 and output feature embeddings of size 256.
The prediction MLPs $\phi$ contain a single hidden layer with a hidden dimension of 1024. 
We apply synchronized batch norm in both MLPs (including the output of $\psi$) following prior work \cite{chen2020simple}.
The classification heads for the temporal self-supervision tasks follow a similar design to $\psi$, except that no batch norm is applied to the output in this case. 

To evaluate models in transfer experiments, we average predictions of multiple temporal and spatial crops. 
Likewise, the features for linear probes and nearest-neighbor retrieval are obtained by averaging multiple crops and standardizing the resulting features using the training set statistics.

\section{Experiments}

\noindent\textbf{Datasets.} As a pre-training dataset we use Kinetics \cite{zisserman2017kinetics} in most of our experiments. 
The dataset contains around 350K training videos categorized into 600 human action classes. 
For transfer experiments we consider UCF101 \cite{soomro2012ucf101} and HMDB51 \cite{hmdb51} which are significantly smaller datasets with human action annotations.
We use these datasets to evaluate the transfer performance of the video branch, both via fine-tuning to action recognition and as fixed feature extractors for video retrieval. 
We evaluate the audio branch of our model on ESC50 \cite{piczak2015esc} in terms of environmental audio classification.\\
\noindent \textbf{Augmented VGG-Sound.}
Finally, we use the test set of VGG-Sound \cite{chen2020vggsound} to evaluate both branches in terms of their robustness to heavy content manipulation for fingerprinting applications. 
Concretely we generate the following four augmented versions of the dataset by applying different types of audio and video transformations (examples in parenthesis):\\
\textbf{1. AugVGG-IP} - "In-Place" manipulations (V: noise, blur, pixelization, emoji overlay; A: noise, clicks).\\
\textbf{2. AugVGG-S} - "Spatial" transformations (V: cropping, padding, rotation; A: pitch shift, reverb, freq. filter).\\
\textbf{3. AugVGG-T} - "Time" transforms (V+A: speed, crops).\\
\textbf{4. AugVGG-C} - "Combined" (one of each type above).\\
We use the AugLy library for the dataset creation \cite{papakipos2022augly}.
% An example augmentation for each set can be seen in Figure~\ref{fig:comp_augvgg}.
For fingerprinting evaluations, we report recall at $k$ for these datasets where queries stem from AugVGG-x and retrievals are computed on the clean test set. 

\subsection{Ablations}

We perform extensive ablation experiments to investigate the influence of the contrastive loss function design, the various temporal self-supervision signals for audio representation learning, and our combined audio-visual model. 
\\

\begin{table}[t]
\centering
\caption{\textbf{Contrastive Loss Design.}
We explore different configurations of the contrastive loss formulation in Eq.~\ref{eq:crl_loss} in combination with temporal SSL when applied to video-video learning (no audio is being used). 
We report nearest-neighbor classifier accuracy on UCF101 and HMDB51 and recall @1 for robust video fingerprinting on VGG-Sound.
}\label{tab:crl_abl}
\resizebox{0.95\linewidth}{!}{%
\begin{tabular}{@{}l@{\hspace{1em}}c@{\hspace{1em}}c@{\hspace{1em}}c@{}}
\toprule
  & {\textbf{UCF101}} & {\textbf{HMDB51}}  & \textbf{AugVGG-C}  \\ 
\textbf{Experiment}   & 1-NN   & 1-NN   & R@1 \\ \midrule
% (a)  w/o NN from $Q_v$ & 77.7 &  63.4 & \underline{51.0} &  33.9 &  60.3 \\ 
(a)  w/o $Q_v$ positives  &    61.5 &   32.5 & \textbf{65.5} \\  
(b)  w/o $Q_v$ negatives  &    \underline{63.9} &    34.0 & \underline{65.1} \\  

(c)  hard negatives   & 63.5 &    33.3  & \textbf{65.5} \\  
% (a) & easy negatives &  77.1 &  64.1  & 49.4 &  33.4 & 47.4 \\  
(d)  easy negatives &     62.9  &  \underline{34.8} &  \underline{65.1}  \\  
(e)  uniform $\omega_j$  &  63.7   &  33.1 &  64.1 \\  
\midrule
Baseline &    \textbf{65.3}   &  \textbf{35.3} & \textbf{65.5} \\ 
\midrule
(f) NNCLR    &  {64.8}   &  {34.2} &  62.8 \\ 
(g) SimCLR    &  53.9   &  29.2 & 61.1 \\ 
(h) SimSiam    &  62.8   &  34.0 & 60.9 \\ 
\bottomrule 
\end{tabular}
}
\label{tab.ablations}
\end{table}

\noindent \textbf{On the Design of the Contrastive Loss.}
We perform experiments with different variants of the general contrastive objective in Equation~\ref{eq:crl_loss} and compare it to some popular existing baselines. 
For faster experimentation, we perform these experiments on video only (we do not use the audio channel here) and pre-train the networks for 40 epochs.
We use an R3D-18 network architecture and perform the temporal reasoning tasks among video clips in the experiments.
We compare the following variants and report results in Table~\ref{tab:crl_abl}:

\noindent \textbf{(a)-(b) Positives and negatives from the memory bank.}
In this case, we remove the nearest neighbors from the memory bank as additional positives (a) or remove the negative sampling from $Q_v$ (b). 
We observe that both positives and negatives from $Q_v$ demonstrate clear benefits, while the positives provide more significant improvements, especially in action retrieval performance. 
% We observe on average worse performance, especially in the nearest neighbor based retrieval experiments.
\\
% \noindent \textbf{(c) Zero-centering the memory bank.}
% By default, we zero-center the memory bank after each update. 
% Our rationale here is to align the output distribution of the projection heads (which is 0-mean due to batch norm) with the distribution of the features in the memory bank.
% Removing this centering from the memory bank hurts performance overall.
% \\
\noindent \textbf{(d)-(e) Difficulty of negatives.}
Instead of sampling negatives starting from the median of nearest neighbors in the memory bank, we start at the 90th percentile for hard negatives in (c) and at the 20th  percentile for easy negatives (d). 
Both variants lead to inferior action retrieval performance, and easy negatives hurt fingerprinting. \\
% We can observe that hard negatives appear to hurt linear probe performance, while easy negatives lead to good linear performance but hurt retrieval. \\
\noindent \textbf{(f) Equal weighting of positives.}
Instead of the cross-view similarity-based weighting of the positives, all five positive examples contribute equally to the loss in this case. 
We observe a drop in the fingerprinting retrieval especially, possibly due to decreased importance of the exact match in the loss. 
This case is similar to the approach in \cite{koohpayegani2021mean}.\\
\noindent \textbf{(g)-(i) Prior approaches.}
We replace our proposed loss with existing prior approaches.
NNCLR \cite{dwibedi2021little} replaces the embedding of one view with its nearest neighbor in the memory bank. 
While this leads to good performance in action retrieval, the performance for fingerprinting suffers. We hypothesize that the lack of the exact match and the lack of additional negatives are the main reason. 
Key differences to SimCLR \cite{chen2020simple} are 1. lack of nearest neighbors, 2. lack of predictor MLP, 3. gradient back-propagation through both views.
SimCLR requires much larger mini-batches to perform well, which is prohibitive on video. 
Finally, SimSiam \cite{chen2020exploring} lacks any negative examples but is otherwise identical to (a). We can again observe the importance of explicit negatives for the fingerprinting use case.

\begin{table}[t]
\centering

\caption{\textbf{Temporal Self-Supervision for Audio Feature Learning.}
We explore how the different temporal self-supervision signals impact the audio representation performance for downstream audio classification on ESC50 and audio fingerprinting on VGG-Sound. The audio encoder is pre-trained with temporal supervision and audio-audio contrastive learning (no RGB frames were used).
}\label{tab:audio_abl}
\resizebox{0.85\linewidth}{!}{%
\begin{tabular}{@{}l@{\hspace{1em}}c@{\hspace{1em}}c@{\hspace{1em}}c@{}}
\toprule
 &  \multicolumn{2}{c}{\textbf{ESC50}} & \textbf{AugVGG-C}  \\ 
\textbf{Ablation} & Linear & 1-NN    & R@1 \\ \midrule

(a)  w/o speed  &  80.4 & 58.4 & 21.1  \\  
(b)  w/o direction  &  79.0 &   56.7 & \underline{21.8}  \\  
(c)  w/o order  & \underline{80.6}  &  \underline{58.5}  & 21.5  \\  
% (d)  w/o audio-audio CLR  &  \underline{81.4} &    {62.9} & 10.6  \\  
(d)  spect.-resize  &  71.0 &   {50.8} &  19.3 \\  
(e)   + rand. STFT-step  &  76.5 &   53.3 &  21.5 \\  
 Baseline & \textbf{82.2} &  \textbf{61.0} &  \textbf{21.9} \\ 
\bottomrule 
\end{tabular}
}
\end{table}

\noindent \textbf{The Benefits of Temporal Self-Supervision for Audio.}
We performed ablation experiments to demonstrate the different temporal learning tasks' effect on audio feature performance. 
We only train the audio branch in these experiments and combine the temporal tasks with an audio-audio contrastive term. 
Networks were again trained for 40 epochs on Kinetics. 
In Table~\ref{tab:audio_abl} (a)-(c), we report the performance of models where each of the three temporal supervision signals is removed. 
We can observe that each task significantly benefits feature performance, especially in downstream audio recognition tasks.
% Interestingly, speed classification appears to hurt audio fingerprinting.
In ablation (d)-(e), the temporal speed transformations are realized by resizing the audio spectrogram instead of subsampling the raw audio signal.
We observe clear performance degradations in these cases, even when randomizing the frame step of the STFT, which could prevent some possible shortcuts due to resizing artifacts. 
% This appears to enable shortcuts further illustrates the importance of randomizing the frame step of the STFT, which can prevent shortcuts and can be considered a form of data augmentation.  
% We also report the results using only the temporal SSL tasks in (e) for completeness. 

\noindent \textbf{Combined Contrastive and Temporal Audio-Visual Learning.}
Finally, we validate our combined audio-visual model through experiments demonstrating the importance of the inclusion (or exclusion) of the different contrastive and temporal objectives and ablate model design variations.
In this set of experiments, we use an R(2+1)D-18 architecture for the video encoder, and we again train the model for 40 epochs.
Table~\ref{tab:mm_abl} shows the results of the following experiments:
\\
\noindent \textbf{(a)-(d) Training Objectives:}
We show the influence of the different contrastive intra- and inter-modal objectives in (a)-(c) and the addition of the temporal reasoning tasks in (d). 
We observe that the cross-modal term brings the most benefit, followed by including the intra-video term. 
Interestingly, the exclusion of the intra-audio term performs better in all cases. 
Finally, note how adding temporal self-supervision to the contrastive objectives provides significant gains across the board. 
\\
\noindent \textbf{(e)-(f) Implementation Details:}
We further illustrate the importance of using temporally aligned positives in the cross-modal contrastive term in (e).
We believe that the model can leverage the temporal audio-visual correspondence to better associate scene events with their sounds. 
% Ablation (f) shows that introducing separate projection and predictor MLPs for the cross-modal video-audio term surprisingly leads to worse performance.  
Finally, in (f), we use only a single memory bank which we feed with the averages of the features from both modalities. Interestingly, this outperforms separate memory banks for fingerprinting and audio recognition.

\begin{table}[t]
\centering
\caption{\textbf{Audio-Visual Model Ablations.}
We perform ablation experiments to demonstrate the influence of the different self-supervised learning signals in our approach (first block) and various implementation details (second block). 
The video encoder is evaluated in transfer to action recognition on UCF101 and HMDB51, and the audio encoder for classification on ESC50.
The fused audio-video feature is used for fingerprinting on VGG-Sound. 
}\label{tab:mm_abl}
\resizebox{\linewidth}{!}{%
\begin{tabular}{@{}l@{\hspace{1em}}c@{\hspace{1em}}c@{\hspace{1em}}c@{\hspace{1em}}c@{}}
\toprule
  &  {\textbf{UCF101}} &  {\textbf{HMDB51}} &  {\textbf{ESC50}} & \textbf{AugVGG-C}  \\ 
\textbf{Experiment}   & 1-NN    & 1-NN    & 1-NN  & R@1 \\ \midrule
%  & no temp + no cross & 74.0 &  55.5 & 42.8 &  27.6  &  \\  
% (a) & BASE & 83.6 &  67.4 & 51.9 &  35.5  & 78.5 & 64.0 \\  
(a) w/o A-V CLR     &  61.0   &  32.2    & 62.9  & 73.8  \\  
(b) w/o V-V CLR   & 61.0   &  33.8   & 67.3 & 69.6 \\  
(c) w/ A-A CLR   & \underline{69.1}   &  37.4   & 68.3 & 78.1 \\  
(d)  w/o temp.-SSL   & 67.5  &  37.4  & 67.4 & 78.6 \\  
\midrule
(e) unaligned A-V   & 68.4  &  37.2   & 68.9 & {78.8} \\  
% (f)  separate heads for video-audio CRL & 81.9  & 64.1  &  50.4 &  33.6 &  \underline{76.3} & 60.5 & 61.6 \\  
(f)  shared $Q$   &  {68.8}    & \underline{38.5}     &  \textbf{69.4} & \textbf{79.3} \\  
% (-)  spectrum transform &  {83.2} &  {68.7} & {52.9}  & \underline{38.5}  & {80.4}  &  {68.6} & {-} \\  

% (a) & consistent reverse+order & 81.7  & 66.9  &  52.7 &  34.9 &  79.8 & 62.8  \\  
% (a) & base  & \underline{82.8}  & \textbf{66.7}  &  \underline{53.4} &  \textbf{37.0} &  \textbf{77.5} & \textbf{64.5}  \\ 
\midrule
Baseline    & \textbf{70.7}    &  \textbf{40.5}   & \underline{69.0} & {78.1}  \\  
% nnprojclose crosssimclr  & {83.0}  & {66.2}  &  {53.8} &  {35.7} &  {78.3} & {64.0}  \\
\bottomrule 
\end{tabular}
}
\label{tab.ablations}
\end{table}

\begin{table*}[t]
    \centering
    \caption{\textbf{Action Recognition on UCF101 and HMDB51 and Audio Classification on ESC50.} We report action recognition accuracy after full fine-tuning and linear probe evaluation. We indicate the pre-training dataset, resolution, the number of frames, iterations (or epochs in brackets), and pre-training data modalities (V=RGB, A=audio).  }
    \label{tab:comparison}
    \resizebox{\linewidth}{!}{%
    \begin{tabular}{@{}l@{\hspace{1em}}c@{\hspace{1em}}c@{\hspace{1em}}c@{\hspace{1em}}c@{\hspace{1em}}c@{\hspace{1em}}c@{\hspace{1em}}c@{\hspace{1em}}c@{\hspace{1em}}c@{\hspace{1em}}c@{\hspace{1em}}c@{}}
    \toprule
      &    &    &  &   &     &       &  \multicolumn{2}{c}{\textbf{UCF101}} & \multicolumn{2}{c}{\textbf{HMDB51}} & \textbf{ESC50} \\ 

    \textbf{Method}      &  \textbf{Dataset} & \textbf{Res.}  &   \textbf{Frames} & \textbf{It. [Ep.]} &  \textbf{Network}  &   \textbf{Mod.}    &   FT    &   Lin. & FT    &   Lin.   &   Lin. \\ \midrule
    TE-CVRL \cite{jenni2021time} &  K400 &  112    &  16  & [200] & R(2+1)D-18 &  V   &    88.2 & &   62.2   \\

    % STS \cite{wang2020self2} & Kinetics  &  224  & 64  & [25]  & S3D-G    &  V+F    &  89.0   & &  62.0  \\
    % CoCLR \cite{han2020self} & K400  &  128  & 32 & [100] & S3D    &  V+F    &  87.9   & 74.5 &  54.6 & 46.1 \\
    % MIL-NCE \cite{miech2020end} &  HowTo100M &  224   &  32 & 400K &   S3D &  V+T   &   \comp{91.3} & \comp{82.7} &  \comp{61.0} & \comp{53.1}  \\
    CVRL \cite{qian2020spatiotemporal} & K600  &  224  & 32  & [800] & R3D-50    &  V    &  \underline{93.4}  & \underline{90.6} &   68.0 & 59.7 \\

    \midrule
    MMV \cite{alayrac2020self} &  AS &  224   &  32  & 500K & R(2+1)D-18 &  V+A   &   \comp{91.5} & \comp{83.9} & \comp{70.1} & \comp{60.0}  \\
    % BraVe \cite{recasens2021broaden} &  AS &  224   &  32  & 620K &  TSM-50 &  V+A   &   \comp{95.6} & \comp{93.4} & \comp{75.3} & \comp{69.1}  \\
    BraVe \cite{recasens2021broaden} &  AS &  224   &  32  & 620K &  R(2+1)D-18 &  V+A   &   \textbf{93.6}  & 90.0 & 70.8 & \underline{63.6}  \\

    \midrule

    AVTS \cite{korbar2018cooperative} &  K400 &  224   &  25  & [90] &  MC3 &  V+A   &   85.8 & & 56.9  & & 76.7 \\
    % AVTS \cite{korbar2018cooperative} &  AudioSet &  224   &  25  & [90] &  MC3 &  V+A   &   \comp{89.0} & & \comp{61.6}   \\

    XDC \cite{alwassel2019self} &  K400 &  224   &  32  & 900K & R(2+1)D-18 &  V+A   &   84.2 & & 47.1  & & 78.5 \\
    % XDC \cite{alwassel2019self} &  AudioSet &  224   &  32  & 2.5M & R(2+1)D-18 &  V+A   &   \comp{91.2} & & \comp{61.0}   \\

    GDT \cite{patrick2020multi} &  K400 &  112   &  32  & [200] & R(2+1)D-18 &  V+A   &   88.7 & &  57.8 & & 78.6  \\

    AVID \cite{Morgado_2021_CVPR} &  K400 &  224   &  32  & [400] &  R(2+1)D-18 &  V+A   &   87.5 & & 60.8  &  & 79.1 \\
    % AVID \cite{Morgado_2021_CVPR} &  AudioSet &  224   &  32  & [400] &  R(2+1)D-18 &  V+A   &   \comp{91.5} & & \comp{64.7}   \\

    \midrule
    \textbf{Ours} &  VGG-S &   112    &  16  & 160K [240] &   R(2+1)D-18 &  V+A   &    90.9 &  86.8 &   70.2  & 55.9 & \textbf{87.9} \\
    
    \textbf{Ours} &  K400 &   112    &  16  & 200K [240] &   R(2+1)D-18 &  V+A   &    91.8 &  88.0 &   71.2  & 58.2 & 84.8 \\
    % \textbf{Ours} &  K600 &   112    &  16  & 210K [160] ? &   R(2+1)D-18 &  V+A   &    92.2 &  89.0 &   72.1  & 60.7 \\
    \textbf{Ours} &  K600 &   112    &  16  & 200K [150] &   R(2+1)D-18 &  V+A   &    {92.2} &  90.3  &   \underline{72.2}  & 62.6 & \underline{86.4} \\

    % \textbf{Ours} &  Kinetics &   112    &  16  & 210K [160] &   R(2+1)D-18 &  V+A   &    91.0 &  87.7 &   70.1  & 59.1 \\

    % \textbf{Ours} &  Kinetics &   224    &  16  & 210K [160] &  R3D-34 &  V+A   & 93.5 &  90.4 &   73.1 & 62.4   \\
    \textbf{Ours} &  K600 &   224    &  16  & 400K [300] &  R3D-34 &  V+A   & \textbf{93.6}  &  \textbf{91.8} &   \textbf{74.6}  &   \textbf{65.8}  & 85.5  \\

    %  VGG-Sound & 170K &  300K  &  90.9  &  86.8    &  70.2   &  55.9 & \textbf{87.9} \\
    %  K400 & 240K & 300K  &  91.8  &  88.0    &  71.2   &  58.2 & 84.8 \\
    %  K600 &  390K & 300K  &  \textbf{92.2}  &  \textbf{89.0}    &   \textbf{72.1}  &  \textbf{60.7} & 86.4  \\

    \bottomrule
    \end{tabular}
    }
\end{table*}

\begin{figure*}[t]
    \centering
    \includegraphics[width=\linewidth]{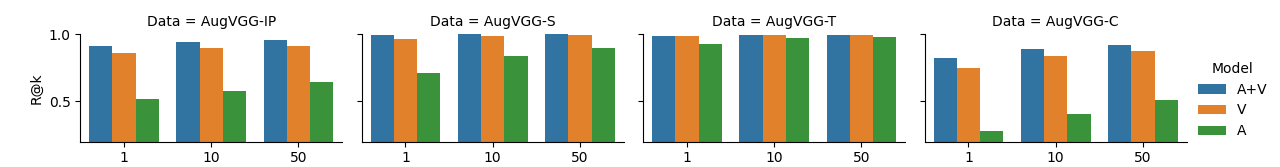}
    \caption{\textbf{Video Fingerprinting Performance.} 
    We report instance retrieval performance under video content manipulation on the different AugVGG variants. 
    We show results using a video only (V), audio only (A), and a joint audio-visual model (A+V). 
    % Examples of the different AugVGG datasets are shown above. 
    }
    \label{fig:comp_augvgg}
\end{figure*}

\subsection{Comparison to Prior Work on Video SSL}

We compare against prior self-supervised video representation learning methods in transfer learning experiments for action recognition and retrieval on UCF101 and HMDB51.
We train and evaluate two different video encoders in these comparisons: 1. a smaller-scale experiment with an R(2+1)D-18 trained at $112\times112$ and 2. a larger-scale experiment with an R3D-34 trained at $224\times224$ resolution.
% Both were pre-trained for 160 epochs using our default training settings. \\

\noindent \textbf{Transfer to Action Recognition and Audio Classification.}
We compare on UCF101 and HMDB51 action recognition and ESC50 audio classification in Table~\ref{tab:comparison}, both with full fine-tuning and linear probes when available.
% We fine-tune the video encoder for X epochs on UCF101 and for Y epochs on HMDB51, using a learning rate of $1\cdot10^{-4}$.
A fair comparison to and among prior works is difficult due to significant differences in pre-training datasets, network architectures, input configurations, and training duration.
We indicate some of these factors that are known to impact performance in the table. 
While there are prior works \cite{recasens2021broaden,qian2020spatiotemporal} reporting comparable performance in some tasks, they either use larger architectures, larger pre-training datasets, train for longer, or a combination of those. 
Our method is more efficient in comparison while still achieving state-of-the-art performance.
Notably, when comparing the most common setting using R(2+1)D-18 trained on Kinetics-400, we outperform the best prior results by +3.1\%, +9.0\%, and +5.7\% on UCF101, HMDB51, and ESC-50 respectively. 
% \textcolor{red}{TODO: expand or highlight some comparisons}

\noindent \textbf{Video Retrieval Performance.}
We compare to the prior state-of-the-art approaches TCLR \cite{dave2021tclr}, GDT \cite{patrick2020multi}, Robust-xID \cite{morgado2021robust}, and TE-CVRL \cite{jenni2021time} in video retrieval benchmarks on UCF101 and HMDB51 in Table~\ref{tab:nn}.
Queries stem from the test set, and retrievals are computed on the training set of the respective dataset. 
A retrieval is assumed correct when the class of query and retrieval agree. 
We report recall at $k$ for different nearest neighbors.
Our model outperforms prior methods by a considerable margin.

\begin{table}[t]
    \centering
    \caption{\textbf{Video Retrieval on UCF101 and HMDB51.} We report recall at $k$ (R@$k$) for $k$-NN video retrieval. All methods use a R(2+1)D-18 network.  }
    \label{tab:nn}
    \resizebox{\linewidth}{!}{%
    \begin{tabular}{@{}l@{\hspace{1em}}c@{\hspace{1em}}c@{\hspace{1em}}c@{\hspace{2em}}c@{\hspace{1em}}c@{\hspace{1em}}c@{\hspace{1em}}c@{\hspace{1em}}c}
    \toprule
         &    \multicolumn{3}{c}{\textbf{UCF101}}  &  \multicolumn{3}{c}{\textbf{HMDB51}}  \\ 

    \textbf{Method}      &  \textbf{R@1}  &  \textbf{R@5}     &   \textbf{R@20}  &   \textbf{R@1}  &  \textbf{R@5}    &   \textbf{R@20}    \\ \midrule
    %Jigsaw \cite{noroozi2016unsupervised} & AlexNet  &  19.7 & 28.5 & 33.5 & 40.0  
    %& - & - & - & -\\ 
    % OPN \cite{lee2017unsupervised} &  AlexNet  &   19.9 & 28.7 & 34.0 & 40.6 
    % & - & - & - & -\\     
    % B\"uchler \etal~\cite{buchler2018improving} &  AlexNet  &   {25.7} & 36.2 & 42.2 &  49.2 & - & - & - & - \\     

    % STS \cite{wang2020self2} &  C3D    &  30.1  & 49.6  & 58.8 & 67.6
    % & 13.9 & 33.3 & 44.7 & 59.5 \\   
    % Pace Pred. \cite{wang2020self} &  C3D    &  31.9  & 49.7  & 59.2 & 68.9
    % & 12.5 & 32.2 & 45.4 & 61.0 \\ 
    % PRP \cite{yao2020video} &  R3D-18    &  22.8  & {38.5}  & {46.7}  & {55.2}
    % & - & - & - & - \\
  
    % VCOP \cite{xu2019self} &  R3D-18    &  14.1  & 30.3  & 40.4 & 51.1
    % & 7.6 & 22.9 & 34.4 & 48.0 \\
    % VCP \cite{luo2020video} &  R3D-18    &  18.6  & 33.6  & 42.5 & 53.5
    % & 7.6 & 24.4 & 36.6 & 53.6 \\
    
    % Var. PSP \cite{cho2020self}  &  R3D-18    &  24.6  & 41.9  & 51.3 & 62.7
    % & 10.3 & 26.6 & 38.8 & 51.6 \\
    
    % PCL \cite{tao2020self}  &  R3D-18    &   40.5  &  59.4  & 68.9  &  77.4
    % &  16.8 &  38.4 &  53.4 & 68.9  \\
                 
    % MemDPC \cite{han2020memory}  &  R3D-18    &  20.2  & 40.4  & 52.4 & 64.7
    % & 7.7 & 25.7 & 40.6 & 57.7 \\
    % Temp.-Trans. \cite{jenni2020video}* &  R3D-18  &  {26.1} & {48.5} & {59.1} & {69.6} 
    % & - & - & - & - \\
    % MCN \cite{lin2021self}* &  R3D-18   &  53.8  & 70.2  & 78.3 & 83.4
    % &  24.1  & 46.8  & 59.7 & 74.2 \\
    
    % SpeedNet \cite{benaim2020speednet}*   &  S3D-G  &  13.0 & 28.1 & 37.5 & 49.5 
    % & - & - & - & -\\
    % CoCLR \cite{han2020self} &  S3D    &  53.3  & 69.4  & 76.6 & 82.0
    % & 23.2 & 43.2 & 53.5 & 65.5 \\

    TCLR     & {56.9}  & {72.2}  &   {84.6}
    & {24.1} & {45.8}  & {75.3} \\

    GDT      & {57.4}  & {73.4}  & {88.1}
    & {25.4} & {51.4}   & {75.0} \\
    
    % TE-CVRL \cite{jenni2021time} &  R(2+1)D    &  {64.3}  & {80.9}  & {86.4} & {90.6} & {29.5} & {55.8} & {68.0} & {78.2} \\

    Robust-xID      &  60.9  & 79.4    & 90.8
    &  30.8  & 55.8    & 79.7 \\
    TE-CVRL      &  \underline{64.2}  & \underline{81.1}   & \underline{92.6}  & \underline{33.1} & \underline{60.8} &   \underline{84.1} \\
    \midrule

    % \textbf{Ours*} &  R(2+1)D-18    &  \underline{76.3}  &  \underline{87.9}  & \underline{92.3} & \underline{95.5}  & \underline{41.6} & \underline{66.0} & \underline{76.3} & \underline{85.6}  \\
    % \textbf{Ours*} &  R3D-34    &   \textbf{83.2}   &   \textbf{92.2}   &  \textbf{95.4}   & \textbf{97.7}
    % &  \textbf{49.1}  &  \textbf{72.6}    &   \textbf{81.5}   &  \textbf{88.4} \\
    
    \textbf{Ours} (R(2+1)D-18)     &  \textbf{80.6}  &  \textbf{90.4}   & \textbf{96.4}  & \textbf{44.9} & \textbf{70.4}  & \textbf{87.6}  \\
    \textbf{Ours} (R3D-34)   &   \textbf{85.2}   &   \textbf{93.0}     & \textbf{97.3}
    &  \textbf{51.3}  &  \textbf{74.3}      &  \textbf{91.4} \\
    % [0.8060846560846561, 0.903968253968254, 0.9373015873015873, 0.964021164021164, 0.9833333333333333]
    %  [0.4489528795811518, 0.7041884816753927, 0.8030104712041884, 0.8763089005235603, 0.9489528795811518]  
    % \textbf{Ours} R3D-34   &   \textbf{85.2}   &   \textbf{93.0}   &  \textbf{95.8}   & \textbf{97.3}
    % &  \textbf{51.3}  &  \textbf{74.3}    &   \textbf{83.9}   &  \textbf{91.4} \\
    % [0.8519597457627118, 0.9303495762711864, 0.9576271186440678, 0.9732521186440678, 0.9899364406779662]
    % [0.5130890052356021, 0.743455497382199, 0.8390052356020943, 0.9142670157068062, 0.9698952879581152]
    \bottomrule
    \end{tabular}
    }
\end{table}

% \noindent \textbf{Audio Classification on ESC50.}
% We also compare our learned audio representations to prior audio-visual video SSL approaches on ESC50 using linear classifiers in Table~\ref{tab:audio_comp}.
% When trained on Kinetics, our approach outperforms prior works on this benchmark. 

% \begin{figure}[t]
%     \centering
%     \includegraphics[width=\linewidth]{figures/full.png}
%     \caption{\textbf{.} 
%     }
%     \label{fig:comp_augvgg}
% \end{figure}

\noindent \textbf{Video Fingerprinting Performance on AugVGG.}
Finally, we report video retrieval performance under video manipulations in Figure~\ref{fig:comp_augvgg}.
We report recall at $k$ for all four datasets and three models: 1. fused audio and video features, 2. video-only, and 3. audio-only.
The fused embedding (concatenation of audio and video features) performs best in all cases, followed by the video model.
Surprisingly, AugVGG-IP with in-place augmentations is most difficult, while performance on AugVGG-S and AugVGG-T is close to perfect. 

\noindent \textbf{Audio-Visual Feature Fusion.}
We explore the fusion of the aural and visual features learned through our approach for downstream video understanding tasks. 
We compare linear probe accuracy for audio, video, and fused features learned on VGG-Sound and Kinetics-600 in Table~\ref{tab:fusion}. 
Interestingly, combining both modalities improves not only the audio-focused VGG-Sound benchmark but also the appearance-focused classification task on Kinetics-600.

% [0.831832627118644, 0.9224046610169492, 0.9536546610169492, 0.9772245762711864, 0.9902012711864406]
% accs_nn: [0.49149214659685864, 0.7264397905759162, 0.8154450261780105, 0.8835078534031413, 0.9633507853403142]
% kNN: [0.7634920634920634, 0.8788359788359789, 0.9230158730158731, 0.9547619047619048, 0.978042328042328]        SVM: 0.8767195767195767
% kNN: [0.4162303664921466, 0.6596858638743456, 0.7630890052356021, 0.856020942408377, 0.9391361256544503]        SVM: 0.5909685863874345

% \begin{table*}[t]
% \centering
% \caption{\textbf{Audio Classification on ESC50.}
% blah
% }\label{tab:audio_comp}
% % \resizebox{\textwidth}{!}{%
% \begin{tabular}{@{}l@{\hspace{1em}}c}
% \toprule
% \textbf{Method} & \textbf{ESC50} \\ \midrule

%  AVTS \cite{korbar2018cooperative} & 76.7  \\ 
%  XDC \cite{alwassel2019self} &  78.5 \\ 
%  GDT \cite{patrick2020multi} &  78.6 \\ 
%  AVID \cite{Morgado_2021_CVPR} &  79.1 \\ 
% \midrule
%  Ours & \textbf{79.9} \\ 
% \bottomrule 
% \end{tabular}
% % }
% \end{table*}

\begin{table}[t]
\centering
\caption{\textbf{Modality Fusion.}
We explore the fusion of our audio-visual features for downstream video classification. 
}\label{tab:fusion}
% \resizebox{\textwidth}{!}{%
\begin{tabular}{@{}l@{\hspace{1em}}c@{\hspace{1em}}c}
\toprule
\textbf{Modalities} & \textbf{VGG-Sound} & \textbf{K600} \\ \midrule

Audio & 39.1  &  15.7 \\ 
Video & \underline{39.7} & \underline{56.8} \\ 
Audio+Video &  \textbf{53.9} & \textbf{58.4} \\ 
\bottomrule 
\end{tabular}
% }
\end{table}

\section{Conclusions}

We introduced a novel method to learn video and audio representations by exploiting temporal and audio-visual self-supervision. 
To learn temporal features, our model learns through time-related pretext tasks, which we extend to the audio domain and the cross-modal setting.
We propose a novel contrastive loss design and a model with both intra- and cross-modal contrastive objectives to learn from the audio-visual correspondence in videos. 
Experiments demonstrate that representations that integrate both temporal and aural features achieve state-of-the-art video classification and retrieval performance.

\bibliography{aaai23}

\end{document}